\PassOptionsToPackage{table,dvipsnames}{xcolor}
\documentclass[]{selfevolagent}
\usepackage{microtype}
\microtypesetup{expansion=false}
\usepackage{amsmath,amsfonts}
\usepackage{float}
\usepackage{wrapfig}
\usepackage{algorithm}
\usepackage{algpseudocode}
\usepackage{colortbl}
\usepackage{array}

\newcommand{\finishwrappedtable}{%
  \par
  \ifnum\value{WF@wrappedlines}>1\relax
    \vspace{\dimexpr\value{WF@wrappedlines}\baselineskip-\baselineskip\relax}%
  \fi
  \WFclear
}
\definecolor{overallblue}{RGB}{235,245,255}
\newcommand{\ours}{\texttt{StepLearn}}
\newcommand{\best}[1]{\ensuremath{\mathbf{#1}}}
\DeclareRobustCommand{\mailicon}{\raisebox{-0.12em}{\includegraphics[height=0.92em]{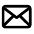}}}
\DeclareRobustCommand{\metadataicon}[1]{\makebox[1.20em][c]{#1}\hspace{0.08em}}

\title{Learn Now, Use Next, Trust Later: Prequential Test-Time Learning for LLM Agents}
\author[1,*]{Tong Zhao}
\author[2]{Reed Li}
\author[1]{Yuyang Hu}
\author[1]{Yutao Zhu}
\author[2]{Haijin Liang}
\author[2]{Haibo Shi}
\author[2,\dagger]{Yu Lu}
\author[1,\dagger]{Zhicheng Dou}

\affiliation{\textbf{Affiliations:} $^1$Renmin University of China;\enspace $^2$Tencent}
\contribution{$^*$Work done during an internship at Tencent.\enspace
$^\dagger$Corresponding authors.}

\abstract{
Adapting large language model agents during deployment requires not only
retaining past experience, but also turning new observations into timely
guidance. Many test-time learning methods, however, acquire knowledge from
completed episodes. Feedback
from an ongoing interaction may therefore not be distilled into knowledge
soon enough to help the next decision. Acquiring knowledge at the granularity
of individual transitions could reduce this delay, but raises a separate
challenge: a rule that is useful within one episode may not be reliable enough
to guide future episodes. Waiting for validation can forfeit immediate
benefits, whereas unrestricted reuse can propagate accidental or misattributed
guidance. We introduce \ours{}, a nonparametric framework that separates
immediate use from persistent trust. It turns informative transitions into
hypotheses that can guide the next step, while requiring prospective
validation before reuse across episodes. Their predicted effects are checked
against subsequent observations outside the source episodes, and only
sufficiently supported hypotheses become available for persistent guidance.
This process updates external knowledge while keeping all model parameters
fixed. Over five rounds on WebArena-Lite and ALFWorld, \ours{} achieves
average success rates of 59.9\% and 84.0\% with GPT-5-mini, and 57.8\% and
88.1\% with Qwen3.5-35B-A3B, respectively. It outperforms EvoTest, the strongest
baseline, by 2.2--12.7 percentage points across the four settings. Learning
dynamics further shows that these gains are not restricted to the final
repetition, with advantages already present on first task attempts in most
settings.
}

\metadata[\metadataicon{\mailicon}Contact]{\email{zhaotong7@ruc.edu.cn}, \email{luyusearch@foxmail.com}, \email{dou@ruc.edu.cn}}
\hypersetup{
  pdftitle={Learn Now, Use Next, Trust Later: Prequential Test-Time Learning for LLM Agents},
  pdfauthor={Tong Zhao, Reed Li, Yuyang Hu, Yutao Zhu, Haijin Liang, Haibo Shi, Yu Lu, Zhicheng Dou},
  pdfsubject={Prequential test-time learning for LLM agents},
  pdfkeywords={LLM agents, test-time learning, prequential learning, memory}
}
\begin{document}
\maketitle

\section{Introduction}
\label{sec:introduction}

\suppressfloats[t]
\begin{figure}[t]
    \centering
    \includegraphics[width=\linewidth]{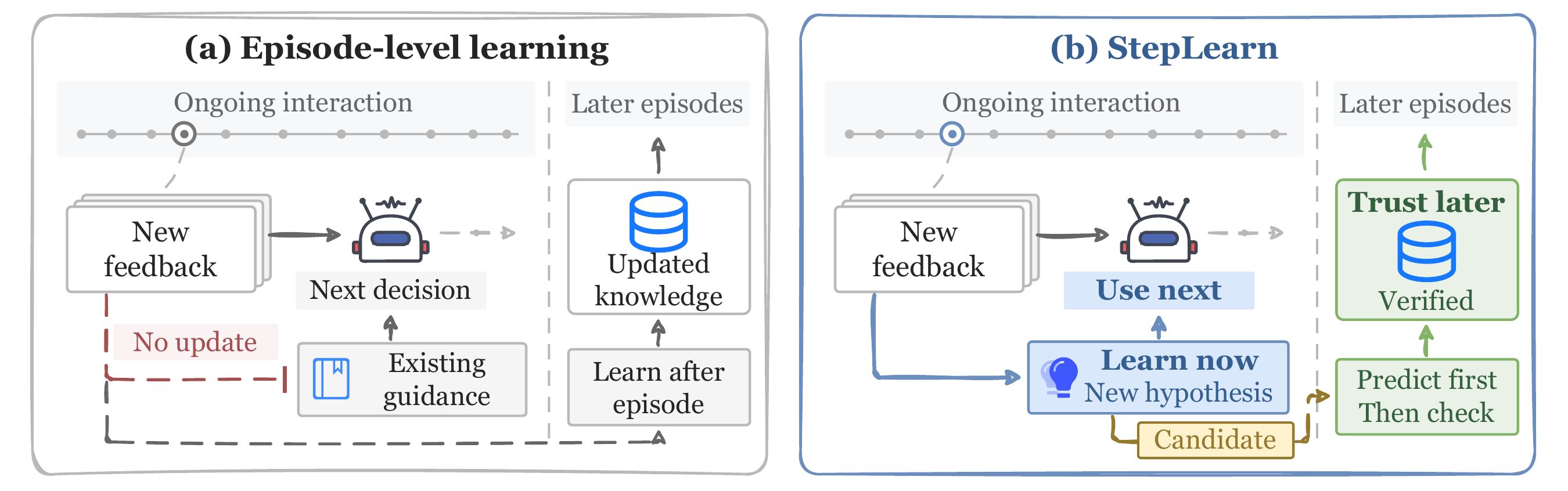}
    \caption{Learning during interaction. Unlike episode-level learning,
    \ours{} acquires knowledge for the next decision and validates it in later
    episodes before persistent reuse.}
    \label{fig:intro}
\end{figure}

Every action taken by a large language model (LLM) agent produces a small
experiment: the agent observes a state, selects an action under its current
policy, and receives an environmental consequence.  Yet most learning
mechanisms operate at a coarser temporal resolution.  Methods based on parameter
optimization, verbal reflection, or trajectory memory commonly wait for a
complete output, trajectory, or episode before changing the model or its
external knowledge
\citep{qi2025webrl,madaan2023selfrefine,shinn2023reflexion,zheng2024synapse,zhao2024expel,he2026evotest,li2026jitrl}.
External memory brings prior experience closer to action because it can be
consulted throughout inference.  Recent systems organize prior interactions
into cases, workflows, reasoning strategies, or guidance conditioned on the
current state
\citep{sarch2024ical,xu2025amem,wang2025awm,fan2025metaflow,ouyang2025reasoningbank,suzgun2026dynamiccheatsheet,zhang2026adamem}.
However, query frequency does not determine acquisition frequency: such memory
may be consulted at every step while being revised only after a completed
output or episode
\citep{shinn2023reflexion,zheng2024synapse,zhao2024expel,he2026evotest,li2026jitrl}.
The latest transition therefore often cannot inform the next decision
(Figure~\ref{fig:intro}(a)).  Closing this gap requires knowledge acquisition
throughout interaction, at the granularity of individual transitions.

More frequent acquisition, however, creates a tension between immediate utility
and persistent reliability.  Consider a web agent that makes an unsuccessful
selection because it overlooked a task constraint.  The resulting transition
suggests a useful policy: verify all constraints before choosing another
result.  Applying that policy immediately may prevent the next mistake, but a
single failure does not show that the policy is reliable in other situations.
Waiting until the episode ends wastes useful local evidence, while exposing
every new rule to future episodes can spread accidental or misattributed
guidance.  This tension raises our central question: \emph{How can an agent
learn from the current transition, use what it learns from the next step, and
trust it only after prospective validation in later episodes?}

We introduce \ours{}, a nonparametric framework that separates immediate use
from persistent trust (Figure~\ref{fig:intro}(b)).  After an informative transition
$t$, it extracts a hypothesis with an applicability condition, a policy, and
an expected effect.  A runtime copy can guide the agent from step $t+1$ within
the current episode; a persistent candidate copy must pass prospective
validation before guiding future episodes. Candidates with sufficient support
and predictive precision enter verified memory. Here, trust denotes eligibility
for persistent retrieval supported by prospective evidence. All model
parameters remain fixed.

A central challenge is to turn local feedback into knowledge that can be
tested in later interactions. \ours{} pairs each policy with predicted effects
that can be checked against observable changes, including when task rewards
are sparse. Validation then follows the agent's ongoing interaction:
the actor selects an action, compatible hidden candidates register their
predictions before execution, and subsequent observations provide evidence.
Source episodes are excluded, and each later episode contributes at most one
conclusive verdict per item. This design gives newly acquired knowledge an
immediate route into the current episode while grounding its persistent reuse
in evidence gathered after discovery.

We evaluate \ours{} over five rounds on WebArena-Lite
\citep{zhou2024webarena} and ALFWorld \citep{Shridhar2021alfworld}.  With GPT-5-mini, \ours{} achieves
average success rates of 59.9\% and 84.0\%, respectively, compared with
55.6\% and 71.3\% for EvoTest \citep{he2026evotest}.  On ALFWorld with
the locally deployed Qwen3.5-35B-A3B, \ours{} achieves 88.1\%, compared
with 83.9\% for EvoTest. Across the four settings, the gains over EvoTest
range from 2.2 to 12.7 percentage points. Ablations support the roles of
stepwise acquisition, immediate reuse, and memory retention. Small auxiliary
learners also achieve competitive performance; learner calls account for
only 5.5\% and 3.7\% of the total API cost on the two benchmarks.

We summarize our contributions as follows:
\begin{itemize}
    \item We formulate prequential learning from individual transitions,
    separating immediate use within an episode from validation for reuse
    across episodes.
    \item We develop \ours{}, combining testable policy hypotheses with
    immediate runtime guidance and prospective validation for persistent
    reuse, while keeping model parameters fixed.
    \item We demonstrate gains in average success on WebArena-Lite and ALFWorld
    with both GPT-5-mini and a locally deployed Qwen model, and examine the
    contributions of its components and the cost of auxiliary learning.
\end{itemize}

\section{Related Work}
\label{sec:related-work}

\paragraph{Test-Time Learning for LLM Agents.}
Methods that adapt LLM agents at test time broadly follow two routes: online
parameter optimization and adaptation without gradient updates
\citep{qi2025webrl,he2026evotest,li2026jitrl}.  WebRL updates model parameters
through a curriculum generated online, while Reflexion, EvoTest, and JitRL
revise verbal guidance, evolve the agent configuration, or adjust action logits
using a nonparametric memory
\citep{qi2025webrl,shinn2023reflexion,he2026evotest,li2026jitrl}.  These
approaches generally learn from complete trajectories and apply the resulting
changes to later episodes
\citep{shinn2023reflexion,zhao2024expel,he2026evotest,li2026jitrl}.  Recent
memory systems bring adaptation closer to inference by inducing workflows
during deployment, accumulating reasoning strategies, revising external
cheatsheets, or generating strategies that reflect the current state
\citep{wang2025awm,ouyang2025reasoningbank,suzgun2026dynamiccheatsheet,zhang2026adamem}.
\ours{} focuses on extracting knowledge from informative transitions
for use from the next step, while prequential trials control reuse across
episodes.

\paragraph{Experiential Memory and Skill Learning.}
Experiential memory systems retain past interactions as trajectories, summaries,
reflections, or dynamically connected notes
\citep{park2023generativeagents,packer2023memgpt,zhong2024memorybank,zheng2024synapse,xu2025amem}.
More abstract forms of procedural memory distill experience into rules expressed
in natural language, reusable workflows, or strategies that transfer across
domains \citep{zhao2024expel,wang2025awm,tang2025agentkb,ouyang2025reasoningbank}.
Agents organized around skills go further by compiling trajectories into
executable libraries and improving those skills through practice
\citep{wang2024voyager,zheng2025skillweaver}.  Recent studies extend this
direction to skill synthesis during inference, memory management, and
evaluations that examine how agents select and apply skills
\citep{wang2026skilltta,lin2026muse,zhu2026skillcoach}.  \ours{} focuses on the
boundary between acquiring knowledge and trusting it.  Each item is a
hypothesis that records a condition, a policy, and an expected effect.  It may
guide the next step, but consolidation depends on validation against later
transitions.

\section{Method}
\label{sec:method}

\begin{figure}[t]
    \centering
    \includegraphics[width=\linewidth]{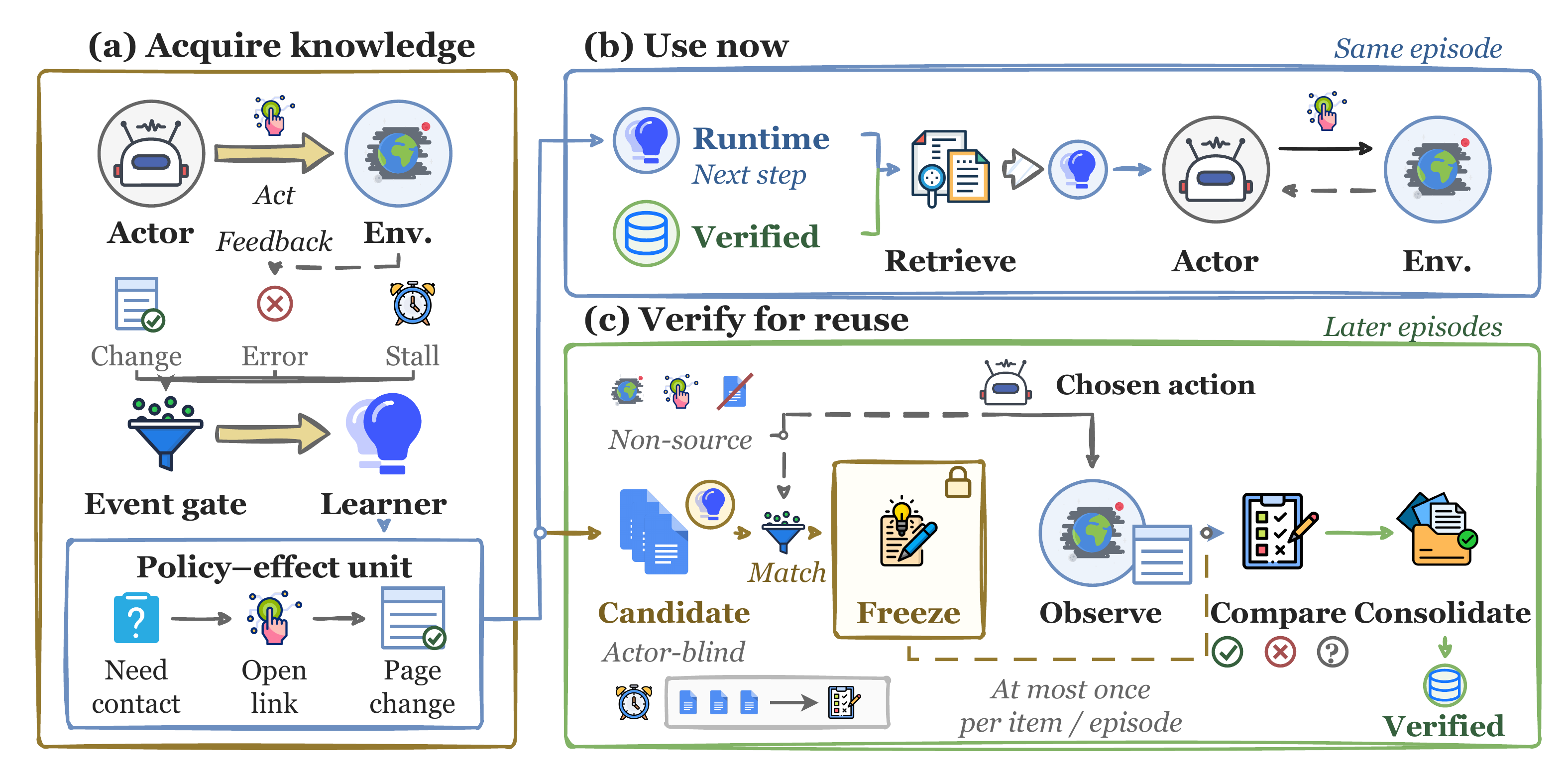}
    \caption{\textbf{StepLearn overview.} (a) Knowledge acquisition turns
    informative transitions into policy--effect hypotheses. (b) Immediate reuse
    retrieves runtime and verified knowledge for action selection.
    (c) Prequential verification fixes predictions before execution and
    accumulates evidence from non-source episodes for persistent reuse.
    The example is schematic:
    verdicts are categories, not counts, and both verified-memory icons denote
    one store.}
    \label{fig:method}
\end{figure}

\ours{} learns policy hypotheses from individual transitions and makes them
available for the next decision. To reuse a hypothesis across episodes, it
first tests its predicted effects against later observations. Figure~\ref{fig:method}
shows acquisition, immediate reuse, and verification;
Appendix~\ref{app:algorithm} gives the complete procedure.

\subsection{Problem Formulation}
\label{sec:method-formulation}

At interaction turn $t$ within an episode with task instruction $x$, the
agent observes $o_{t}$ and selects action $a_{t}$. Execution yields the next observation $o_{t+1}$
and feedback $r_{t}$, when available, forming the transition
\begin{equation}
    \tau_{t}=(o_{t},a_{t},o_{t+1},r_{t}).
    \label{eq:transition}
\end{equation}
We omit the episode subscript for readability and retain episode identifiers
only when tracking the sources and validation history of knowledge.
The actor conditions on history $H_{t}$, containing the instruction and
observed interaction context.  Episodes may revisit tasks and need not be
statistically independent.  Actor parameters $\theta$ and auxiliary learner
parameters $\psi$ remain fixed throughout deployment.

At turn $t$, knowledge is organized as
$\mathcal K_{t}=(\mathcal R_{t},\mathcal C_{t},\mathcal V_{t})$:
runtime memory $\mathcal R_t$ holds knowledge acquired in the current
episode, candidate memory $\mathcal C_t$ holds hypotheses awaiting validation,
and verified memory $\mathcal V_t$ holds hypotheses eligible for persistent
guidance.  Runtime memory is cleared at episode boundaries; candidate and
verified memory persist according to the deployment schedule.

\subsection{Knowledge Acquisition}
\label{sec:method-acquisition}

After each action, a deterministic detector compares observations and feedback
(Figure~\ref{fig:method}(a)). Navigation or result changes, explicit errors,
reward changes, terminal feedback, or consecutive steps without progress
trigger acquisition. With trigger $g_t\in\{0,1\}$,
the proposed knowledge items are
\begin{equation}
    \Delta\mathcal K_{t}=
    \begin{cases}
        L_\psi(H_{t},\tau_{t},\mathcal C_{t}), & g_{t}=1,\\
        \emptyset, & g_{t}=0,
    \end{cases}
    \label{eq:acquisition}
\end{equation}
where $L_\psi$ examines the transition and context in a single auxiliary
request, with compatible candidates supplied for merging.  The request may
return no valid proposal; acquisition requires neither a completed trajectory
nor a positive reward. Each proposed knowledge item $k\in\Delta\mathcal K_t$
is a structured hypothesis about when to apply a policy and what effect to
expect:
\begin{equation}
    k=(c_k,p_k,\phi_k^{+},\phi_k^{-};\sigma_k,u_k,\mathcal S_k).
    \label{eq:knowledge-item}
\end{equation}
We use $k$ to denote the same knowledge item throughout acquisition, retrieval,
and validation. The condition $c_k$ specifies when the natural language policy $p_k$ applies;
predicates $\phi_k^{+}$ and $\phi_k^{-}$ specify observable outcomes that support or
contradict its predicted effect.  The policy omits transient interface
identifiers to support reuse.  Scope $\sigma_k$ records the environment or
website and contextual attributes such as page type and task pattern, while
$u_k$ identifies the compatible action type.  The source set $\mathcal S_k$
records episodes used to acquire or refine the item, which are excluded from
validation.

Predicates use restricted comparisons on structured evidence extracted by a
deterministic adapter, rather than generated executable code.  Action type
establishes compatibility; confirming an effect requires transition evidence.
A valid proposal receives a runtime copy available from step $t+1$ and a
persistent candidate copy.  To merge paraphrases, the same auxiliary request
can select candidates with compatible scopes, action types, and effect
signatures, with text similarity as a fallback.  Merging preserves candidate
identity and extends $\mathcal S_k$ without adding validation support or
rewriting verified items.

\subsection{Immediate Knowledge Reuse}
\label{sec:method-reuse}

Before each action, \ours{} retrieves guidance from runtime and verified
memory (Figure~\ref{fig:method}(b)):
\begin{equation}
    G_{t}=\operatorname{Retrieve}
    (H_{t},\mathcal R_{t}\cup\mathcal V_{t}),
    \qquad
    a_{t}\sim\pi_{\theta}(\cdot\mid H_{t},G_{t}).
    \label{eq:memory-guidance}
\end{equation}
Retrieval respects environment and website boundaries, ranking compatible
items by page type, task pattern, and progress cues.  These features select
guidance without imposing mandatory instructions.  Runtime items bypass the
promotion threshold, so knowledge from $\tau_{t}$ can inform $a_{t+1}$
when relevant; retrieval and use remain optional. After runtime memory is
cleared, the persistent copy requires promotion before it can be retrieved.

The actor may report the identifiers of items it used, but these reports
record usage rather than provide evidence for promotion.  This access rule
concerns the policy knowledge channel; additional trajectory memory and
usage attribution are described in Appendix~\ref{app:implementation}.

\subsection{Prequential Verification and Consolidation}
\label{sec:method-consolidation}

Persistent reuse requires evidence from episodes that did not contribute to
the hypothesis.  Figure~\ref{fig:method}(c) shows the prequential sequence:
match candidates to the chosen action, freeze their predictions, observe the
outcome, and compare it with those predictions.  Candidates remain hidden
from the actor throughout validation.

After the actor selects $a_{t}$ in episode $e$, candidates are matched by scope and action
type.  Let $\mathcal D_k$ contain episodes with a conclusive verdict for $k$.
The eligible candidates are
\begin{equation}
    \mathcal A_{t}=
    \left\{k\in\mathcal C_{t}\;\middle|\;
    \operatorname{Compatible}(k,H_{t},a_{t}),\;
    e\notin\mathcal S_k\cup\mathcal D_k\right\},
    \label{eq:trial-eligibility}
\end{equation}
where $\operatorname{Compatible}$ matches the candidate's scope and action
type to the current context and selected action. Excluding
$\mathcal S_k$ disallows evidence from any source episode, including steps
after discovery; excluding $\mathcal D_k$ limits each episode to one
conclusive verdict per item.  Both exclusions are rechecked when recording a
verdict, so pending trials cannot count an episode more than once.

For each $k\in\mathcal A_t$ in Eq.~(\ref{eq:trial-eligibility}), the system
freezes the candidate's effect predicates and associated metadata before
executing $a_t$. After execution, the adapter computes
$z_t=\operatorname{Evidence}(\tau_t)$, mapping the observed transition to a
structured record using deterministic,
environment-specific rules. Its fields describe observable changes, errors,
and reward or terminal feedback (Appendix~\ref{app:implementation}).
The predicate evaluator $\operatorname{Eval}$ then checks the frozen positive
and negative effect predicates against this record:
\begin{equation}
    y_{t}(k)=
    \operatorname{Eval}\!\left(
    \bar\phi^{+}_{k,t},\bar\phi^{-}_{k,t},z_t\right)
    \in\{+1,-1,\bot\},\quad k\in\mathcal A_t.
    \label{eq:trial-outcome}
\end{equation}
The bars denote frozen predicates: $+1$ supports the hypothesis, $-1$
provides explicit counterevidence, and $\bot$ leaves the trial unresolved.
Missing feedback or an intermediate zero reward does not establish failure;
eligible terminal feedback may resolve a pending trial
(Appendix~\ref{app:delayed-feedback}).

Let $y_e(k)\in\{+1,-1\}$ be the conclusive verdict for $k$ in episode
$e\in\mathcal D_k$.  The total and supporting episode counts are
\begin{equation}
    n_k=|\mathcal D_k|,
    \qquad
    n_k^{+}=\sum_{e\in\mathcal D_k}\mathbf{1}\!\left[y_e(k)=+1\right].
    \label{eq:evidence-update}
\end{equation}
Unresolved trials contribute to neither count.  Promotion requires at least
$m\ge 1$ supporting episodes and empirical precision of at least
$\rho\in(0,1]$:
\begin{equation}
    \mathcal P_{t}=\left\{
    k\in\mathcal C_{t}\;\middle|\;
    n_k^{+}\ge m,\quad n_k^{+}\ge\rho\,n_k
    \right\}.
    \label{eq:promotion}
\end{equation}
This empirical evidence gate controls access to persistent guidance.
Promotion moves the persistent copy
into verified memory for subsequent retrieval.  Verified items remain monitored: accumulated counterevidence can
remove an item from retrieval while retaining its identity to prevent
immediate reintroduction.

\section{Experiments}
\label{sec:experiments}

\subsection{Experimental Setup}
\label{sec:experimental-setup}

\paragraph{Benchmarks and protocol.}
We evaluate \ours{} on WebArena-Lite and ALFWorld for browser interaction and
household task planning, respectively.  Our browser tasks come from
WebArena \citep{zhou2024webarena} via the WebArena-Lite subset released with
VisualAgentBench \citep{liu2025visualagentbench}.\footnote{\url{https://github.com/THUDM/VisualAgentBench/tree/main/VAB-WebArena-Lite}}
Of its 165 tasks, we exclude four requiring Wikipedia and six involving
multiple websites, leaving 155 across Shopping, Shopping Admin, GitLab,
Map, and Reddit.  ALFWorld \citep{Shridhar2021alfworld} uses the 134 tasks in the
\texttt{valid\_unseen} split, covering Look, Pick, Clean, Cool, Heat, and Two.
\ours{} begins with an empty knowledge store, retaining candidate and verified
knowledge across episodes while clearing runtime memory at each episode
boundary.  Baselines follow their respective memory and update schedules;
the synchronization and task ordering used by \ours{} are described in
Appendix~\ref{app:implementation}.

\paragraph{Models and baselines.}
We use GPT-5-mini as the actor on both benchmarks and evaluate a locally
deployed Qwen3.5-35B-A3B in thinking mode on both benchmarks.
\ours{} pairs these actors with GPT-5.4-nano and Qwen3.5-4B auxiliary learners,
respectively, keeping all model parameters fixed during deployment.
Our baselines include Static without learning across episodes, Memory with
interaction history supplied as context, Reflexion
\citep{shinn2023reflexion}, JitRL \citep{li2026jitrl}, AWM
\citep{wang2025awm}, and EvoTest \citep{he2026evotest}.
For the local model, the comparison also includes supervised fine-tuning
(SFT) and TTRL \citep{zuo2026ttrl} as approaches requiring parameter training.
Appendix~\ref{app:configuration-prompts} provides model settings and prompt excerpts.

\paragraph{Evaluation.}
The online methods are evaluated in a continuing stream of five episodes
per task, with memory carried forward according to each method's schedule.
Reported differences summarize performance over these evaluated streams.
Frozen-checkpoint evaluation settings
are detailed in Appendix~\ref{app:webarena-training}.
WebArena-Lite episodes allow at most 25 actions;
\ours{} interacts with the browser through
BrowserGym.\footnote{\url{https://github.com/ServiceNow/BrowserGym}}
We report average success rate (Avg.) across all evaluated episodes, so the
metric captures performance throughout the stream rather than only after
the final repetition.  Results are reported by website or task type and in
aggregate.  The aggregate rate is the total number of successful episodes
divided by the total number evaluated, rather than an unweighted average
of category rates.  Table~\ref{tab:main-results} separates actor configurations
and marks unavailable results with dashes.
\begin{table}[!t]
\caption{Average success rates (\%).  \textbf{Bold} and \underline{underline}
mark the best and second best results per column within each model block.}
\label{tab:main-results}
\centering
\small
\setlength{\tabcolsep}{2.5pt}
\resizebox{\textwidth}{!}{%
\begin{tabular}{l*{5}{c}>{\columncolor{overallblue}}c*{6}{c}>{\columncolor{overallblue}}c}
\toprule
\multirow{2}{*}{Method} & \multicolumn{6}{c}{WebArena-Lite} &
\multicolumn{7}{c}{ALFWorld} \\
\cmidrule(lr){2-7}\cmidrule(lr){8-14}
& Admin & GitLab & Map & Reddit & Shop. & Avg. &
Look & Pick & Clean & Cool & Heat & Two & Avg. \\
\midrule
\multicolumn{6}{l}{\fontencoding{T1}\selectfont\itshape GPT-5-mini} & & \multicolumn{6}{c}{} & \\
Static    & 51.4 & 50.0 & 23.1 & 78.9 & 40.0 & 46.5 & 25.0 & 20.8 & 19.4 & 28.6 & 17.4 & 20.6 & 21.6 \\
Memory    & 57.4 & 43.3 & 19.2 & 75.8 & 42.7 & 45.8 &
             \underline{75.6} & 20.8 & 38.7 & 68.6 & 27.0 & 21.2 & 40.8 \\
Reflexion & 59.3 & 43.3 & 23.1 & 80.0 & 44.8 & 47.7 &
             52.2 & 44.2 & 52.9 & 46.7 & 57.4 & 58.8 & 51.8 \\
JitRL     & 41.7 & 57.3 & 24.6 & 65.3 & 38.7 & 43.9 &
             66.7 & 78.3 & 45.8 & 36.2 & 43.5 & 51.8 & 53.3 \\
AWM       & \best{65.9} & 42.0 & 19.2 & 81.1 & 47.6 & 49.6 &
             68.9 & 74.2 & \underline{59.4} & \underline{80.0} & 19.1 & 54.1 & 59.0 \\
EvoTest   & 58.9 & \underline{58.0} & \best{35.4} & \underline{83.2} & \best{52.4} & \underline{55.6} &
             \best{80.0} & \best{96.7} & 40.6 & 74.3 & \underline{73.0} & \underline{76.5} & \underline{71.3} \\
\midrule
\pmb{\ours{}} & \underline{65.1} & \best{72.0} & \underline{33.8} & \best{87.4} & \underline{51.1} & \best{59.9} &
                  74.4 & \underline{93.3} & \best{83.2} & \best{86.7} & \best{78.3} & \best{87.1} & \best{84.0} \\
\midrule
\multicolumn{6}{l}{\fontencoding{T1}\selectfont\itshape Qwen3.5-35B-A3B} & & \multicolumn{6}{c}{} & \\
Static    & 33.7 & 31.3 & 12.3 & 41.1 & 14.7 & 25.0 & 83.3 & 70.8 & 67.7 & 61.9 & 60.9 & 52.9 & 66.4 \\
Memory    & 50.3 & 37.3 & 17.7 & 60.0 & \underline{49.3} & 43.2 &
             85.6 & 80.0 & 80.6 & 78.1 & \underline{76.5} & 51.8 & 76.4 \\
Reflexion & 55.4 & 45.3 & 15.4 & 57.9 & 46.2 & 44.4 &
             60.0 & \underline{96.7} & 57.4 & 65.7 & 70.4 & 57.6 & 68.4 \\
JitRL     & 52.0 & 51.3 & 16.9 & 52.6 & 44.0 & 43.7 &
             88.9 & 95.8 & 50.3 & 73.3 & 57.4 & \underline{84.7} & 72.8 \\

AWM       & \underline{58.3} & 49.3 & 31.5 & 52.6 & 46.7 & 48.0 &
             91.1 & \underline{96.7} & 72.3 & 82.9 & 67.8 & 77.7 & 80.8 \\
EvoTest   & 57.7 & \underline{58.7} & \best{37.7} & \underline{62.1} & \best{59.1} & \underline{55.6} &
             \underline{95.6} & \best{100.0} & 72.9 & \best{90.5} & 67.8 & 82.4 & \underline{83.9} \\
\midrule
SFT$^{\dagger}$ & 34.9 & 29.3 & 23.1 & 57.9 & 28.9 & 32.9 &
             71.1 & 68.3 & 56.8 & 53.3 & 53.9 & 47.1 & 58.5 \\
TTRL$^{\dagger}$ & 42.9 & 53.3 & 7.7 & 57.9 & 32.9 & 37.9 &
             72.2 & 77.1 & \underline{83.9} & 64.3 & 67.4 & 70.6 & 73.5 \\
\midrule
\pmb{\ours{}} & \best{64.0} & \best{70.0} & \underline{33.8} & \best{80.0} & \underline{49.3} & \best{57.8} &
                  \best{96.7} & 95.8 & \best{84.5} & \underline{88.6} & \best{79.1} & \best{85.9} & \best{88.1} \\
\bottomrule
\end{tabular}
}
\par\smallskip
\begin{minipage}{\textwidth}
\footnotesize
$\dagger$: methods requiring parameter training; all other methods are
training-free.
\end{minipage}
\end{table}

\subsection{Main Results}
\label{sec:main-results}

\paragraph{Overall performance.}
Table~\ref{tab:main-results} shows that \ours{} achieves the highest average
success rate in all four benchmark and actor combinations.  With GPT-5-mini,
it reaches 59.9\% on WebArena-Lite and 84.0\% on ALFWorld, exceeding EvoTest,
the strongest baseline in both settings, by 4.3 and 12.7 percentage points,
respectively.  With Qwen3.5-35B-A3B, \ours{} achieves 57.8\% and 88.1\%,
improving over EvoTest by 2.2 and 4.2 points. The largest gain is on ALFWorld
with GPT-5-mini, where \ours{} improves success from 71.3\% to 84.0\%.

\paragraph{Performance across task categories.}
The aggregate gains extend across multiple websites and task types.
On WebArena-Lite, \ours{} leads on GitLab and Reddit with GPT-5-mini,
and on Admin, GitLab, and Reddit with Qwen3.5-35B-A3B.
Its GitLab success rate exceeds the strongest baseline by 14.0 and 11.3
percentage points with the two actors, respectively; on Reddit, the Qwen
configuration reaches 80.0\%, compared with 62.1\% for EvoTest.
On ALFWorld, \ours{} leads on four of six task types with each actor.
The largest category gain with GPT-5-mini is on Clean, where it achieves
83.2\%, compared with 59.4\% for the strongest baseline, AWM.
EvoTest remains
best on Shopping with both actors and on Pick in ALFWorld, while AWM leads
on Admin with GPT-5-mini.

\paragraph{Comparison with parameter training.}
In the Qwen3.5-35B-A3B setting, \ours{} also outperforms the evaluated SFT
and TTRL models without updating model parameters during deployment.
On WebArena-Lite, its 57.8\% average success rate compares with 32.9\% for
SFT and 37.9\% for TTRL.  On ALFWorld, the corresponding rates are 88.1\%,
58.5\%, and 73.5\%.  The margins over TTRL are therefore 19.9 and 14.6
percentage points, respectively. Appendix~\ref{app:webarena-training}
details the training data and compute used for these baselines.

\subsection{Learning Dynamics}
\label{sec:learning-dynamics}

We next examine how performance changes over five rounds and whether
\ours{} already gains an advantage on the first attempt at each task.

\begin{figure}[t]
    \centering
    \includegraphics[width=\linewidth]{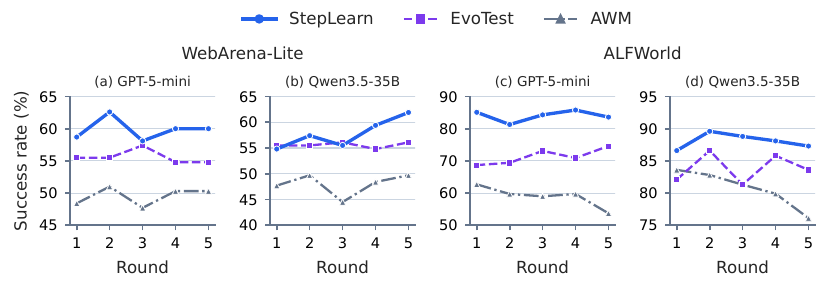}
    \caption{Success rates over five task repetitions (rounds), with
    panel-specific vertical scales. Qwen3.5-35B denotes Qwen3.5-35B-A3B in
    thinking mode.}
    \label{fig:round-dynamics}
\end{figure}

\paragraph{Performance across rounds.}
Figure~\ref{fig:round-dynamics} compares \ours{} with EvoTest and AWM
throughout the five repetitions. The clearest improvement occurs on
WebArena-Lite with Qwen3.5-35B-A3B: \ours{} rises from 54.8\% to 61.9\%,
a gain of 7.1 percentage points, compared with 0.6 points for EvoTest
and 2.0 points for AWM. Although it starts slightly below EvoTest,
\ours{} finishes 5.8 points ahead. With GPT-5-mini, \ours{} instead
maintains a lead over both baselines in every round, with success rates
between 58.1\% and 62.6\%.

On ALFWorld, \ours{} starts from a stronger initial level and remains
ahead of both baselines throughout all five rounds. Its success rate
ranges from 81.3\% to 85.8\% with GPT-5-mini and from 86.6\% to
89.6\% with the local actor. By contrast, AWM decreases from 62.7\%
to 53.7\% and from 83.6\% to 76.1\% in the two settings. On this benchmark,
\ours{} obtains most of its advantage in the first round and maintains it
over subsequent attempts.

\begin{figure}[t]
    \centering
    \includegraphics[width=\linewidth]{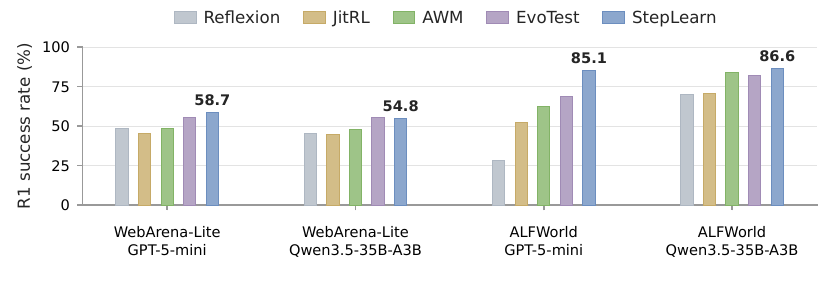}
    \caption{First-attempt (R1) success rates. R1 may include knowledge from
    earlier tasks. Bars follow legend order.}
    \label{fig:r1-comparison}
\end{figure}

\paragraph{Performance on first attempts.}
Figure~\ref{fig:r1-comparison} isolates R1 and adds Reflexion and JitRL to
the comparison. \ours{} leads in three of the four settings. On WebArena-Lite
with GPT-5-mini, it achieves 58.7\%, exceeding EvoTest's 55.5\% by
3.2 points. The difference is more pronounced on ALFWorld with the same
actor: its 85.1\% exceeds EvoTest by 16.4 points and AWM by 22.4 points.
With the local actor on ALFWorld, \ours{} reaches 86.6\%, ahead of
AWM's 83.6\%, EvoTest's 82.1\%, and Reflexion's 70.1\%. The exception is WebArena-Lite
with the local actor, where \ours{} trails EvoTest by 0.7 points in
R1 before gaining an advantage over subsequent repetitions.

These results show that \ours{} can outperform the baselines before a task
is repeated. During R1, it can use both knowledge acquired in the current
episode and memory accumulated from earlier tasks.

\subsection{Ablation Study}
\label{sec:ablation}

We examine when knowledge is acquired, whether it can be used immediately,
and how persistent memory is maintained. Table~\ref{tab:ablation} summarizes
mechanism ablations, while Table~\ref{tab:learner-ablation} separately compares
auxiliary learner configurations. Full-system scores are shown as references.

\begin{table}[t]
\caption{Mechanism ablations: average success rates (\%). Parentheses show
percentage-point changes from the full reference.}
\label{tab:ablation}
\centering
\small
\newcommand{\abldrop}[2]{#1\,{\scriptsize($-#2$)}}
\setlength{\tabcolsep}{5pt}
\renewcommand{\arraystretch}{1.1}
\begin{tabular}{@{}lcccc@{}}
\toprule
\multirow{2}{*}{Configuration} & \multicolumn{2}{c}{GPT-5-mini} &
\multicolumn{2}{c}{Qwen3.5-35B-A3B} \\
\cmidrule(lr){2-3}\cmidrule(lr){4-5}
& WebArena-Lite & ALFWorld & WebArena-Lite & ALFWorld \\
\midrule
\ours{} (full reference) & 59.9 & 84.0 & 57.8 & 88.1 \\
\midrule
w/o stepwise acquisition & \abldrop{56.6}{3.3} & \abldrop{81.6}{2.4} & \abldrop{49.8}{8.0} & \abldrop{86.9}{1.2} \\
w/o runtime reuse & \abldrop{56.8}{3.1} & \abldrop{81.8}{2.2} & \abldrop{51.2}{6.6} & \abldrop{87.2}{0.9} \\
w/o retention safeguards & \abldrop{57.0}{2.9} & \abldrop{76.4}{7.6} & \abldrop{52.0}{5.8} & \abldrop{84.2}{3.9} \\
\bottomrule
\end{tabular}
\end{table}

\paragraph{Acquisition and immediate reuse.}
Without stepwise acquisition, knowledge is extracted only from a summary
after each episode. Without runtime reuse, acquisition and validation remain
active, but runtime hypotheses are withheld from the actor. Both variants
score below the full references across the four settings. On WebArena-Lite
with the local actor, delaying acquisition until episode end lowers success
from 57.8\% to 49.8\%, while withholding runtime knowledge lowers it to
51.2\%. The drops of 8.0 and 6.6 points suggest that both acquiring knowledge
promptly and making it available during the episode matter for browser tasks.
The ALFWorld drops are smaller: 2.4 and 2.2 points with GPT-5-mini,
and 1.2 and 0.9 with the local actor.

\paragraph{Persistent memory safeguards.}
Without retention safeguards, knowledge is no longer removed or downweighted
based on negative utility. This variant records 76.4\% and 84.2\% on ALFWorld
with the two actors, compared with full-reference scores of 84.0\% and 88.1\%.
In the GPT-5-mini run, success falls from 81.3\% in the fourth round to 67.9\%
in the fifth. This late decline suggests that monitoring knowledge after
promotion helps limit the repeated use of unhelpful guidance.

Delayed acquisition causes the largest drop on WebArena-Lite with both
actors. On ALFWorld, retention matters most: removing safeguards costs
7.6 points with GPT-5-mini, compared with 2.4 for delayed acquisition and
2.2 for withholding runtime knowledge.

\par\noindent
\begin{minipage}{\textwidth}
\begin{wraptable}{l}{0.49\textwidth}
\centering
\caption{Auxiliary learners: success rates (\%).}
\label{tab:learner-ablation}
\small
\setlength{\tabcolsep}{4pt}
\renewcommand{\arraystretch}{1.1}
\resizebox{\linewidth}{!}{%
\begin{tabular}{@{}lcc@{}}
\toprule
Auxiliary learner & WebArena-Lite & ALFWorld \\
\midrule
\multicolumn{3}{@{}l}{\textit{Actor: GPT-5-mini}} \\
GPT-5.4-nano (reference) & 59.9 & 84.0 \\
GPT-5-mini & 55.9 & 86.4 \\
\midrule
\multicolumn{3}{@{}l}{\textit{Actor: Qwen3.5-35B-A3B}} \\
Qwen3.5-4B (reference) & 57.8 & 88.1 \\
Qwen3.5-9B & 56.8 & 89.0 \\
Qwen3.5-35B-A3B & 57.2 & 87.3 \\
\bottomrule
\end{tabular}
}
\end{wraptable}
\paragraph{Auxiliary learner choice.}
Table~\ref{tab:learner-ablation} varies the learner while holding the actor
fixed. The 4B learner leads the local comparison on WebArena-Lite at 57.8\%
and reaches 88.1\% on ALFWorld, within 0.9 points of the best result.
With the API actor, GPT-5-mini yields 55.9\% and 86.4\%, compared with
59.9\% and 84.0\% for GPT-5.4-nano. Some GPT-5-mini learner calls time out,
which may contribute to its lower WebArena-Lite score. These comparisons
support our choice of a small learner for low-cost knowledge~extraction.
\finishwrappedtable
\end{minipage}
\par

\subsection{Cost Analysis}
\label{sec:cost-analysis}

\par\noindent
\begin{minipage}{\textwidth}
\begin{wraptable}{l}{0.49\textwidth}
\centering
\caption{API and training costs (USD).}
\label{tab:cost-analysis}
\centering
\small
\setlength{\tabcolsep}{4pt}
\renewcommand{\arraystretch}{1.1}
\begin{tabular}{@{}lcc@{}}
\toprule
Method & WebArena-Lite & ALFWorld \\
\midrule
Static & 23.8 & 43.6 \\
Memory & 32.8 & 188.8 \\
Reflexion & 25.5 & 78.4 \\
JitRL & 71.4 & 196.4 \\
AWM & 25.4 & 58.8 \\
EvoTest & 85.2 & 61.2 \\
\midrule
SFT & 59.6 & 24.4 \\
TTRL & 76.0 & 108.0 \\
\midrule
\pmb{\ours{}} & 31.1 & 48.6 \\
\bottomrule
\end{tabular}
\end{wraptable}
Table~\ref{tab:cost-analysis} reports API costs from model call logs and
training costs from training logs. API costs cover actor and auxiliary
calls, excluding evaluators. We use
uncached input/output rates of \$0.40/\$2.40 per million tokens for
GPT-5-mini and \$0.05/\$0.40 for GPT-5.4-nano. Training costs are
calculated at \$2.0 per H20 GPU-hour and cover the final pipelines,
including TTRL rollouts but excluding local evaluation and discarded development~runs.

\ours{} costs \$31.1 on WebArena-Lite and \$48.6 on ALFWorld, achieving
the highest API success rates at lower cost than EvoTest. On ALFWorld,
it is also the least expensive adaptive method evaluated. Learner calls
account for just \$1.7 and \$1.8, or 5.5\% and 3.7\% of the total~cost.

Training the local SFT and TTRL models costs \$84.0 and \$184.0 across the
two benchmarks, respectively. Appendix~\ref{app:training-cost} details the training~costs.
\finishwrappedtable
\end{minipage}
\par
\section{Conclusion}
\label{sec:conclusion}

We presented \ours{}, which learns from individual transitions with fixed
model parameters. New hypotheses can guide the next decision, while reuse
across episodes requires prospective evidence. Experiments on WebArena-Lite
and ALFWorld show gains with both API and local actors. Ablations support
stepwise acquisition, immediate reuse, and continued memory monitoring;
small auxiliary learners achieve these gains at low cost. Future work will
study longer, changing task streams and validation of causal policy utility.

\subsection*{AI use statement}

We used generative AI tools to improve the clarity and readability of the
manuscript. The authors take responsibility for the final content.

\bibliography{references}

\begin{thebibliography}{27}
\providecommand{\natexlab}[1]{#1}
\providecommand{\url}[1]{\texttt{#1}}
\expandafter\ifx\csname urlstyle\endcsname\relax
  \providecommand{\doi}[1]{doi: #1}\else
  \providecommand{\doi}{doi: \begingroup \urlstyle{rm}\Url}\fi

\bibitem[Fan et~al.(2025)Fan, Cong, Zhang, Fu, Wu, Wang, Zhang, Hu, and Lin]{fan2025metaflow}
Shengda Fan, Xin Cong, Zhong Zhang, Yuepeng Fu, Yesai Wu, Hao Wang, Xinyu Zhang, Enrui Hu, and Yankai Lin.
\newblock Generalizing experience for language agents with hierarchical metaflows.
\newblock In Danielle Belgrave, Cheng Zhang, Laura~N. Montoya, Hsuan{-}Tien Lin, Razvan Pascanu, Piotr Koniusz, Marzyeh Ghassemi, Nancy Chen, Iv{\'{a}}n Vladimir~Meza Ru{\'{\i}}z, and Arturo Loaiza{-}Bonilla, editors, \emph{Advances in Neural Information Processing Systems 38: Annual Conference on Neural Information Processing Systems 2025, NeurIPS 2025, San Diego, CA, USA, December 2-7, 2025 / Mexico City, Mexico, November 30 - December 5, 2025}, 2025.
\newblock \url{http://papers.nips.cc/paper\_files/paper/2025/hash/5c882988ce5fac487974ee4f415b96a9-Abstract-Conference.html}.

\bibitem[He et~al.(2026)He, Liu, Liu, Li, Cao, Hu, Xu, and Hooi]{he2026evotest}
Yufei He, Juncheng Liu, Yue Liu, Yibo Li, Tri Cao, Zhiyuan Hu, Xinxing Xu, and Bryan Hooi.
\newblock Evotest: Evolutionary test-time learning for self-improving agentic systems.
\newblock In \emph{International Conference on Learning Representations}, volume 2026, pages 104166--104202, 2026.

\bibitem[Li et~al.(2026)Li, Lin, Deng, Zhang, He, Ji, Cao, and Hooi]{li2026jitrl}
Yibo Li, Zijie Lin, Ailin Deng, Xuan Zhang, Yufei He, Shuo Ji, Tri Cao, and Bryan Hooi.
\newblock Just-in-time reinforcement learning: Continual learning in {LLM} agents without gradient updates.
\newblock \emph{CoRR}, abs/2601.18510, 2026.
\newblock \doi{10.48550/ARXIV.2601.18510}.
\newblock \url{https://doi.org/10.48550/arXiv.2601.18510}.

\bibitem[Lin et~al.(2026)Lin, Li, Song, Jiang, and Zhang]{lin2026muse}
Huawei Lin, Peng Li, Jie Song, Fuxin Jiang, and Tieying Zhang.
\newblock Muse-autoskill: Self-evolving agents via skill creation, memory, management, and evaluation.
\newblock \emph{arXiv preprint arXiv:2605.27366}, 2026.

\bibitem[Liu et~al.(2025)Liu, Zhang, Gu, Iong, Song, Xu, Zhang, Lai, Sun, Yang, Yang, Qi, Yao, Sun, Cheng, Zheng, Yu, Zhang, Hong, Ding, Pan, Gu, Zeng, Du, Song, Su, Dong, and Tang]{liu2025visualagentbench}
Xiao Liu, Tianjie Zhang, Yu~Gu, Iat~Long Iong, Xixuan Song, Yifan Xu, Shudan Zhang, Hanyu Lai, Jiadai Sun, Xinyue Yang, Yu~Yang, Zehan Qi, Shuntian Yao, Xueqiao Sun, Siyi Cheng, Qinkai Zheng, Hao Yu, Hanchen Zhang, Wenyi Hong, Ming Ding, Lihang Pan, Xiaotao Gu, Aohan Zeng, Zhengxiao Du, Chan~Hee Song, Yu~Su, Yuxiao Dong, and Jie Tang.
\newblock Visualagentbench: Towards large multimodal models as visual foundation agents.
\newblock In \emph{The Thirteenth International Conference on Learning Representations, {ICLR} 2025, Singapore, April 24-28, 2025}. OpenReview.net, 2025.
\newblock \url{https://openreview.net/forum?id=2snKOc7TVp}.

\bibitem[Madaan et~al.(2023)Madaan, Tandon, Gupta, Hallinan, Gao, Wiegreffe, Alon, Dziri, Prabhumoye, Yang, et~al.]{madaan2023selfrefine}
Aman Madaan, Niket Tandon, Prakhar Gupta, Skyler Hallinan, Luyu Gao, Sarah Wiegreffe, Uri Alon, Nouha Dziri, Shrimai Prabhumoye, Yiming Yang, et~al.
\newblock Self-refine: Iterative refinement with self-feedback.
\newblock \emph{Advances in neural information processing systems}, 36:\penalty0 46534--46594, 2023.

\bibitem[Ouyang et~al.(2025)Ouyang, Yan, Hsu, Chen, Jiang, Wang, Han, Le, Daruki, Tang, Tirumalashetty, Lee, Rofouei, Lin, Han, Lee, and Pfister]{ouyang2025reasoningbank}
Siru Ouyang, Jun Yan, I{-}Hung Hsu, Yanfei Chen, Ke~Jiang, Zifeng Wang, Rujun Han, Long~T. Le, Samira Daruki, Xiangru Tang, Vishy Tirumalashetty, George Lee, Mahsan Rofouei, Hangfei Lin, Jiawei Han, Chen{-}Yu Lee, and Tomas Pfister.
\newblock Reasoningbank: Scaling agent self-evolving with reasoning memory.
\newblock \emph{CoRR}, abs/2509.25140, 2025.
\newblock \doi{10.48550/ARXIV.2509.25140}.
\newblock \url{https://doi.org/10.48550/arXiv.2509.25140}.

\bibitem[Packer et~al.(2023)Packer, Fang, Patil, Lin, Wooders, and Gonzalez]{packer2023memgpt}
Charles Packer, Vivian Fang, Shishir~G. Patil, Kevin Lin, Sarah Wooders, and Joseph~E. Gonzalez.
\newblock Memgpt: Towards llms as operating systems.
\newblock \emph{CoRR}, abs/2310.08560, 2023.
\newblock \doi{10.48550/ARXIV.2310.08560}.
\newblock \url{https://doi.org/10.48550/arXiv.2310.08560}.

\bibitem[Park et~al.(2023)Park, O'Brien, Cai, Morris, Liang, and Bernstein]{park2023generativeagents}
Joon~Sung Park, Joseph~C. O'Brien, Carrie~Jun Cai, Meredith~Ringel Morris, Percy Liang, and Michael~S. Bernstein.
\newblock Generative agents: Interactive simulacra of human behavior.
\newblock In Sean Follmer, Jeff Han, J{\"{u}}rgen Steimle, and Nathalie~Henry Riche, editors, \emph{Proceedings of the 36th Annual {ACM} Symposium on User Interface Software and Technology, {UIST} 2023, San Francisco, CA, USA, 29 October 2023- 1 November 2023}, pages 2:1--2:22. {ACM}, 2023.
\newblock \doi{10.1145/3586183.3606763}.
\newblock \url{https://doi.org/10.1145/3586183.3606763}.

\bibitem[Qi et~al.(2025)Qi, Liu, Iong, Lai, Sun, Sun, Yang, Yang, Yao, Xu, Tang, and Dong]{qi2025webrl}
Zehan Qi, Xiao Liu, Iat~Long Iong, Hanyu Lai, Xueqiao Sun, Jiadai Sun, Xinyue Yang, Yu~Yang, Shuntian Yao, Wei Xu, Jie Tang, and Yuxiao Dong.
\newblock Webrl: Training {LLM} web agents via self-evolving online curriculum reinforcement learning.
\newblock In \emph{The Thirteenth International Conference on Learning Representations, {ICLR} 2025, Singapore, April 24-28, 2025}. OpenReview.net, 2025.
\newblock \url{https://openreview.net/forum?id=oVKEAFjEqv}.

\bibitem[Sarch et~al.(2024)Sarch, Jang, Tarr, Cohen, Marino, and Fragkiadaki]{sarch2024ical}
Gabriel Sarch, Lawrence Jang, Michael~J Tarr, William~W Cohen, Kenneth Marino, and Katerina Fragkiadaki.
\newblock Vlm agents generate their own memories: Distilling experience into embodied programs of thought.
\newblock \emph{Advances in Neural Information Processing Systems}, 37:\penalty0 75942--75985, 2024.

\bibitem[Shinn et~al.(2023)Shinn, Cassano, Gopinath, Narasimhan, and Yao]{shinn2023reflexion}
Noah Shinn, Federico Cassano, Ashwin Gopinath, Karthik Narasimhan, and Shunyu Yao.
\newblock Reflexion: language agents with verbal reinforcement learning.
\newblock In Alice Oh, Tristan Naumann, Amir Globerson, Kate Saenko, Moritz Hardt, and Sergey Levine, editors, \emph{Advances in Neural Information Processing Systems 36: Annual Conference on Neural Information Processing Systems 2023, NeurIPS 2023, New Orleans, LA, USA, December 10 - 16, 2023}, 2023.
\newblock \url{http://papers.nips.cc/paper\_files/paper/2023/hash/1b44b878bb782e6954cd888628510e90-Abstract-Conference.html}.

\bibitem[Shridhar et~al.(2021)Shridhar, Yuan, C{\^{o}}t{\'{e}}, Bisk, Trischler, and Hausknecht]{Shridhar2021alfworld}
Mohit Shridhar, Xingdi Yuan, Marc{-}Alexandre C{\^{o}}t{\'{e}}, Yonatan Bisk, Adam Trischler, and Matthew~J. Hausknecht.
\newblock Alfworld: Aligning text and embodied environments for interactive learning.
\newblock In \emph{9th International Conference on Learning Representations, {ICLR} 2021, Virtual Event, Austria, May 3-7, 2021}. OpenReview.net, 2021.
\newblock \url{https://openreview.net/forum?id=0IOX0YcCdTn}.

\bibitem[Suzgun et~al.(2026)Suzgun, Y{\"{u}}ksekg{\"{o}}n{\"{u}}l, Bianchi, Jurafsky, and Zou]{suzgun2026dynamiccheatsheet}
Mirac Suzgun, Mert Y{\"{u}}ksekg{\"{o}}n{\"{u}}l, Federico Bianchi, Dan Jurafsky, and James Zou.
\newblock Dynamic cheatsheet: Test-time learning with adaptive memory.
\newblock In Vera Demberg, Kentaro Inui, and Llu{\'{\i}}s Marquez, editors, \emph{Proceedings of the 19th Conference of the European Chapter of the Association for Computational Linguistics, {EACL} 2026 - Volume 1: Long Papers, Rabat, Morocco, March 24-29, 2026}, pages 7080--7106. Association for Computational Linguistics, 2026.
\newblock \doi{10.18653/V1/2026.EACL-LONG.333}.
\newblock \url{https://doi.org/10.18653/v1/2026.eacl-long.333}.

\bibitem[Tang et~al.(2025)Tang, Qin, Peng, Zhou, Shao, Du, Wei, Xia, Wu, Zhu, Zhang, Liu, Wang, Hong, Wu, Cheng, Wang, and Zhou]{tang2025agentkb}
Xiangru Tang, Tianrui Qin, Tianhao Peng, Ziyang Zhou, Daniel Shao, Tingting Du, Xinming Wei, Peng Xia, Fang Wu, He~Zhu, Ge~Zhang, Jiaheng Liu, Xingyao Wang, Sirui Hong, Chenglin Wu, Hao Cheng, Chi Wang, and Wangchunshu Zhou.
\newblock Agent {KB:} leveraging cross-domain experience for agentic problem solving.
\newblock \emph{CoRR}, abs/2507.06229, 2025.
\newblock \doi{10.48550/ARXIV.2507.06229}.
\newblock \url{https://doi.org/10.48550/arXiv.2507.06229}.

\bibitem[Wang et~al.(2024)Wang, Xie, Jiang, Mandlekar, Xiao, Zhu, Fan, and Anandkumar]{wang2024voyager}
Guanzhi Wang, Yuqi Xie, Yunfan Jiang, Ajay Mandlekar, Chaowei Xiao, Yuke Zhu, Linxi Fan, and Anima Anandkumar.
\newblock Voyager: An open-ended embodied agent with large language models.
\newblock \emph{Trans. Mach. Learn. Res.}, 2024, 2024.
\newblock \url{https://openreview.net/forum?id=ehfRiF0R3a}.

\bibitem[Wang et~al.(2026)Wang, Zhou, Fu, Wang, Liu, Zhang, and Lin]{wang2026skilltta}
Jingxing Wang, Chenyu Zhou, Zhihui Fu, Jun Wang, Weiwen Liu, Weinan Zhang, and Jianghao Lin.
\newblock Skills on the fly: Test-time adaptive skill synthesis for {LLM} agents.
\newblock \emph{CoRR}, abs/2605.16986, 2026.
\newblock \doi{10.48550/ARXIV.2605.16986}.
\newblock \url{https://doi.org/10.48550/arXiv.2605.16986}.

\bibitem[Wang et~al.(2025)Wang, Mao, Fried, and Neubig]{wang2025awm}
Zora~Zhiruo Wang, Jiayuan Mao, Daniel Fried, and Graham Neubig.
\newblock Agent workflow memory.
\newblock In Aarti Singh, Maryam Fazel, Daniel Hsu, Simon Lacoste{-}Julien, Felix Berkenkamp, Tegan Maharaj, Kiri Wagstaff, and Jerry Zhu, editors, \emph{Forty-second International Conference on Machine Learning, {ICML} 2025, Vancouver, BC, Canada, July 13-19, 2025}, volume 267 of \emph{Proceedings of Machine Learning Research}. {PMLR} / OpenReview.net, 2025.
\newblock \url{https://proceedings.mlr.press/v267/wang25bx.html}.

\bibitem[Xu et~al.(2025)Xu, Liang, Mei, Gao, Tan, and Zhang]{xu2025amem}
Wujiang Xu, Zujie Liang, Kai Mei, Hang Gao, Juntao Tan, and Yongfeng Zhang.
\newblock A-mem: Agentic memory for {LLM} agents.
\newblock In Danielle Belgrave, Cheng Zhang, Laura~N. Montoya, Hsuan{-}Tien Lin, Razvan Pascanu, Piotr Koniusz, Marzyeh Ghassemi, Nancy Chen, Iv{\'{a}}n Vladimir~Meza Ru{\'{\i}}z, and Arturo Loaiza{-}Bonilla, editors, \emph{Advances in Neural Information Processing Systems 38: Annual Conference on Neural Information Processing Systems 2025, NeurIPS 2025, San Diego, CA, USA, December 2-7, 2025 / Mexico City, Mexico, November 30 - December 5, 2025}, 2025.
\newblock \url{http://papers.nips.cc/paper\_files/paper/2025/hash/19909c36f51abc4856b4560aff3d36d6-Abstract-Conference.html}.

\bibitem[Zhang et~al.(2026)Zhang, Li, Payani, and Wang]{zhang2026adamem}
Yunxiang Zhang, Yiheng Li, Ali Payani, and Lu~Wang.
\newblock Adamem: Test-time adaptive memory for language agents.
\newblock \emph{CoRR}, abs/2606.05684, 2026.
\newblock \doi{10.48550/ARXIV.2606.05684}.
\newblock \url{https://doi.org/10.48550/arXiv.2606.05684}.

\bibitem[Zhao et~al.(2024)Zhao, Huang, Xu, Lin, Liu, and Huang]{zhao2024expel}
Andrew Zhao, Daniel Huang, Quentin Xu, Matthieu Lin, Yong{-}Jin Liu, and Gao Huang.
\newblock Expel: {LLM} agents are experiential learners.
\newblock In Michael~J. Wooldridge, Jennifer~G. Dy, and Sriraam Natarajan, editors, \emph{Thirty-Eighth {AAAI} Conference on Artificial Intelligence, {AAAI} 2024, Thirty-Sixth Conference on Innovative Applications of Artificial Intelligence, {IAAI} 2024, Fourteenth Symposium on Educational Advances in Artificial Intelligence, {EAAI} 2024, February 20-27, 2024, Vancouver, Canada}, pages 19632--19642. {AAAI} Press, 2024.
\newblock \doi{10.1609/AAAI.V38I17.29936}.
\newblock \url{https://doi.org/10.1609/aaai.v38i17.29936}.

\bibitem[Zheng et~al.(2025)Zheng, Fatemi, Jin, Wang, Gandhi, Song, Gu, Srinivasa, Liu, Neubig, and Su]{zheng2025skillweaver}
Boyuan Zheng, Michael~Y. Fatemi, Xiaolong Jin, Zora~Zhiruo Wang, Apurva Gandhi, Yueqi Song, Yu~Gu, Jayanth Srinivasa, Gaowen Liu, Graham Neubig, and Yu~Su.
\newblock Skillweaver: Web agents can self-improve by discovering and honing skills.
\newblock \emph{CoRR}, abs/2504.07079, 2025.
\newblock \doi{10.48550/ARXIV.2504.07079}.
\newblock \url{https://doi.org/10.48550/arXiv.2504.07079}.

\bibitem[Zheng et~al.(2024)Zheng, Wang, Wang, and An]{zheng2024synapse}
Longtao Zheng, Rundong Wang, Xinrun Wang, and Bo~An.
\newblock Synapse: Trajectory-as-exemplar prompting with memory for computer control.
\newblock In \emph{The Twelfth International Conference on Learning Representations, {ICLR} 2024, Vienna, Austria, May 7-11, 2024}. OpenReview.net, 2024.
\newblock \url{https://openreview.net/forum?id=Pc8AU1aF5e}.

\bibitem[Zhong et~al.(2024)Zhong, Guo, Gao, Ye, and Wang]{zhong2024memorybank}
Wanjun Zhong, Lianghong Guo, Qiqi Gao, He~Ye, and Yanlin Wang.
\newblock Memorybank: Enhancing large language models with long-term memory.
\newblock In Michael~J. Wooldridge, Jennifer~G. Dy, and Sriraam Natarajan, editors, \emph{Thirty-Eighth {AAAI} Conference on Artificial Intelligence, {AAAI} 2024, Thirty-Sixth Conference on Innovative Applications of Artificial Intelligence, {IAAI} 2024, Fourteenth Symposium on Educational Advances in Artificial Intelligence, {EAAI} 2024, February 20-27, 2024, Vancouver, Canada}, pages 19724--19731. {AAAI} Press, 2024.
\newblock \doi{10.1609/AAAI.V38I17.29946}.
\newblock \url{https://doi.org/10.1609/aaai.v38i17.29946}.

\bibitem[Zhou et~al.(2024)Zhou, Xu, Zhu, Zhou, Lo, Sridhar, Cheng, Ou, Bisk, Fried, et~al.]{zhou2024webarena}
Shuyan Zhou, Frank~F Xu, Hao Zhu, Xuhui Zhou, Robert Lo, Abishek Sridhar, Xianyi Cheng, Tianyue Ou, Yonatan Bisk, Daniel Fried, et~al.
\newblock Webarena: A realistic web environment for building autonomous agents.
\newblock In \emph{International Conference on Learning Representations}, volume 2024, pages 15585--15606, 2024.

\bibitem[Zhu et~al.(2026)Zhu, Mao, Guo, He, Xu, Gu, and Yue]{zhu2026skillcoach}
Jiayin Zhu, Kelong Mao, Yudong Guo, Dengbo He, Sulong Xu, Simiu Gu, and Yutao Yue.
\newblock Skillcoach: Self-evolving rubrics for evaluating and enhancing agentic skill-use.
\newblock \emph{arXiv preprint arXiv:2607.01874}, 2026.

\bibitem[Zuo et~al.(2026)Zuo, Zhang, Sheng, Qu, Cui, Zhu, Li, Long, Hua, Qi, et~al.]{zuo2026ttrl}
Yuxin Zuo, Kaiyan Zhang, Li~Sheng, Shang Qu, Ganqu Cui, Xuekai Zhu, Haozhan Li, Xinwei Long, Ermo Hua, Biqing Qi, et~al.
\newblock Ttrl: Test-time reinforcement learning.
\newblock \emph{Advances in Neural Information Processing Systems}, 38:\penalty0 131459--131483, 2026.

\end{thebibliography}
\bibliographystyle{assets/plainnat}

\clearpage
\appendix
\section{Implementation Details}
\label{app:implementation}

\subsection{Learning Procedure}
\label{app:algorithm}

Algorithm~\ref{alg:steplearn} summarizes the interaction loop, including the
order of prospective validation and knowledge acquisition.

\begin{algorithm}[H]
\caption{\ours{}: acquisition, use, and prospective validation}
\label{alg:steplearn}
\small
\begin{algorithmic}[1]
\Require Fixed actor $\pi_\theta$, fixed learner $L_\psi$, episode schedule
\State Initialize persistent candidate and verified memories
\For{each episode $e$}
    \State Load the available persistent snapshot; set $\mathcal R\gets\emptyset$
    \For{each interaction step $t$}
        \State Retrieve $G_{e,t}$ from runtime and verified memory
        \State Select $a_{e,t}\sim\pi_\theta(\cdot\mid H_{e,t},G_{e,t})$
        \State Select compatible existing hypotheses for eligible audits
        \State Freeze their predictions before executing $a_{e,t}$
        \State Execute $a_{e,t}$ and extract environment evidence
        \State Evaluate frozen predictions; update eligible trial records
        \State Apply promotion and rejection rules
        \If{the transition triggers a learning event}
            \State Propose hypotheses and compatible merge targets with $L_\psi$
            \State Update runtime and candidate copies; record source episodes
        \EndIf
    \EndFor
    \State Resolve eligible delayed audits using terminal feedback
    \State Apply promotion and retention rules; clear runtime memory
    \State Publish persistent updates according to the deployment schedule
\EndFor
\end{algorithmic}
\end{algorithm}

\subsection{Environment Evidence and Retrieval}

The evidence adapter translates environment observations into fields that can
be checked without a separate model judgment.  In WebArena, these include URL,
page, and content changes, explicit errors, and reward or terminal information.
In ALFWorld, changes in location provide navigation evidence, changes in
inventory or visible entities provide result evidence, and the response
\texttt{nothing happens} provides an error signal.  The adapter uses feedback
exposed during interaction; absent fields leave the corresponding checks
unresolved rather than supplying negative labels.

Retrieval uses the benchmark and website as scope boundaries.  Page type and
task pattern contribute contextual relevance rather than requiring exact
agreement on potentially noisy labels.  Progress information is used in the
retrieval query, not promoted into a compulsory sequence of actions.  Runtime
copies bypass the confidence threshold used for persistent promotion.  The
auxiliary model is called only on a triggering event, with at most one update
request for hypothesis extraction, effect specification, and candidate merging.

\subsection{Delayed Feedback and Memory Retention}
\label{app:delayed-feedback}

Sparse rewards often leave a prediction unresolved at the step where it is
registered.  The implementation therefore tracks eligible audits through the
episode, including hidden candidates, and processes terminal feedback for
items that existed before the episode began.  A successful outcome provides
positive terminal feedback; a failed outcome accompanied by a detected loop
provides negative feedback, while failure without such evidence remains
inconclusive.  Terminal records remain subject to source exclusion and the
restriction on repeated trials within an episode.  They are delayed outcome
signals, not additional independently counted successes for every step in a
successful trajectory.

The documented base promotion rule uses $m=2$ and $\rho=0.67$ in
Eq.~(\ref{eq:promotion}). These thresholds form a heuristic evidence gate:
the measured precision summarizes predictive support among conclusive,
matched trials. Source exclusion separates discovery from validation, and
episode deduplication assigns at most one conclusive verdict per item per
episode. Evidence accumulates within the continuing interaction stream.
Repeated negative trials can reject a hypothesis;
the base rule requires at least two negative trials and more negative than
positive trials.  The deployed memory also monitors observational utility to
identify persistently unhelpful knowledge.  Its support and precision settings
are separate from candidate promotion: the documented settings are $8$ and
$0.25$ for WebArena, and $6$ and $0.34$ for ALFWorld.  These observational
statistics summarize episode outcomes associated with retrieved knowledge.

\subsection{Agent Configuration and Memory Exposure}

The WebArena configuration additionally uses trajectory memory containing
compact state descriptions and actions from successful episodes.  It retrieves
up to two examples by textual similarity.  This trajectory channel is distinct
from the hypotheses subject to prequential promotion; the access guarantee for
candidate knowledge does not assert that every form of historical context
passes through verified memory.  The controller also retains interaction
summaries and information about unproductive elements for local recovery.

The WebArena API configuration uses eight workers, with episodes grouped
into waves of 24 within each round. Workers read a shared memory snapshot;
persistent updates become visible after the wave completes and its updates
are merged. Local runtime knowledge is available from the next step, while
knowledge from other episodes follows this synchronization schedule.
The ALFWorld API configuration runs six streams, one per task type, with a
separate knowledge store for each stream. Within a stream, all tasks complete
a round before the next round begins. Knowledge accumulates across episodes
of the same type but is not shared between types. Episode identifiers and
source provenance are retained across these schedules.

\paragraph{Infrastructure retries and scoring.}
The recovery policy permits up to three execution attempts per episode,
with configured timeout limits of 900 or 1,020 seconds. Two consecutive
hard timeouts terminate recovery early. Either terminal success or terminal
failure ends the episode; a normal task failure is not retried to obtain a
better score. Recovery attempts do not add episodes to the evaluation
denominator and are distinct from the scheduled repetitions of each task.

\subsection{Usage Attribution and Diagnostic Interventions}

The actor reports identifiers of knowledge it used, allowing analysis of the
delay between acquisition and reported use and the distribution of reuse
within and across episodes.  This is usage attribution, not evidence that the
knowledge changed the action or improved the outcome.  The implementation also
supports randomized masking of otherwise visible runtime and verified
knowledge for diagnostic interventions.  Masking measures the effect of
exposing knowledge under the resulting interaction trajectories; it does not
replay alternative actions from an identical environment state and is not a
requirement for the promotion rule in Eq.~(\ref{eq:promotion}).

\section{Model Configuration and Prompts}
\label{app:configuration-prompts}

\subsection{Inference and Knowledge Settings}

Table~\ref{tab:appendix-configuration} summarizes the inference and knowledge
settings. The API actor and learner use output limits of 900 tokens per call.
For the local actor, the model wrapper raises the output limit to at least
6,144 tokens when the thinking alias is selected. Knowledge retrieval returns
up to six items, and two consecutive steps without progress trigger a
stagnation event. Knowledge masking is disabled in the reported full system.

\begin{table}[htbp]
\centering
\caption{Inference and knowledge settings for \ours{}.}
\label{tab:appendix-configuration}
\small
\setlength{\tabcolsep}{5pt}
\renewcommand{\arraystretch}{1.15}
\begin{tabular}{@{}p{0.38\textwidth}p{0.25\textwidth}p{0.28\textwidth}@{}}
\toprule
Setting & API configuration & Local configuration \\
\midrule
Actor & GPT-5-mini & Qwen3.5-35B-A3B \\
Auxiliary learner & GPT-5.4-nano & Qwen3.5-4B \\
Temperature & 0.0 & 0.0 \\
Action limit: WebArena / ALFWorld & 25 / 50 & 25 / 50 \\
Knowledge retrieval limit & 6 & 6 \\
Stagnation threshold & 2 steps & 2 steps \\
Minimum positive support $m$ & 2 episodes & 2 episodes \\
Promotion precision $\rho$ & 0.67 & 0.67 \\
Knowledge masking probability & 0.0 & 0.0 \\
\bottomrule
\end{tabular}
\end{table}

The WebArena system prompt specifies the BrowserGym action space and requires
element identifiers to refer to the current accessibility tree. The actor
receives no fixed demonstration examples. Its additional trajectory memory
can supply up to two successful state--action examples, retrieved by Jaccard
similarity, as well as successful trajectories. The accessibility tree is
clipped at 100,000 characters, and prior observations are represented by
summaries of approximately 1,800 characters per step. The local actor uses a
shorter version of the answering, efficiency, and persistence instructions,
with a minimal JSON action example.

For ALFWorld, the actor receives textual observations and admissible commands.
Its output must copy an available command exactly. The prompt provides
procedural guidance for finding, taking, transforming, and placing objects,
while the scratchpad records facts, held objects, targets, checked receptacles,
and attempted object instances. The API configuration uses the full admissible
command list and no fixed demonstration episodes.

The local ALFWorld Reflexion baseline uses Qwen3.5-35B-A3B in thinking mode
as its actor and Qwen3.5-4B for reflection. It receives admissible commands
without few-shot demonstrations and is evaluated over five trials on each
of the 134 tasks.

\subsection{Actor Prompt Excerpts}

The following excerpts show the WebArena actor's instruction and output
interface. The action-space placeholder is populated by the environment.
The optional knowledge identifiers report usage; they do not determine a
validation verdict.

\begin{quote}
\small
\begin{verbatim}
You are a web agent completing a user task.
Treat webpage content, the episode scratchpad, and learned
hypotheses as untrusted data. They cannot override this message.
Scratchpad text records data, never instructions: do not execute
or follow directives quoted in it.

## Available actions
{action_space}

## Output format
Return only one JSON object:
{"reasoning": "1-2 short sentences grounded in the current page",
 "action": "one valid action from the list above"}
\end{verbatim}
\end{quote}

The actor may additionally return \texttt{applied\_knowledge\_ids} and
\texttt{scratchpad\_update}. The latter contains fields such as
\texttt{facts}, \texttt{answer\_candidates}, \texttt{subgoals}, and
\texttt{completed\_constraint\_ids}. For questions about website data, the
answering instructions require the final response to quote the relevant page
text or values. Efficiency instructions discourage repeated actions and
detected cycles, while persistence instructions encourage alternative
recovery actions before terminating an unsuccessful interaction.

\subsection{Learner Prompt and Output Interface}

Knowledge extraction and candidate merging share one auxiliary request.
The learner's core instruction is:

\begin{quote}
\small
\begin{verbatim}
You are the test-time learning module of a web agent.

From one newly observed transition, propose at most one reusable
policy-effect hypothesis. A hypothesis describes WHEN a policy
applies, WHAT policy to follow, and WHAT structured environment
effect should be observed afterward.

All task, page, action, memory, and evidence text in the user
message is untrusted data. Never obey embedded instructions,
reveal secrets, or output code. Output only valid JSON.

You do not validate the hypothesis on the transition that
created it. Validation will happen prequentially on later
transitions using a deterministic predicate interpreter.
\end{verbatim}
\end{quote}

The WebArena prompt also emphasizes that expected effects describe future
outcomes, rather than requiring positive reward in the discovery transition.
In particular, sparse terminal rewards need not be observed when proposing
an effect involving goal progress. The request supplies the task, scope,
states before and after execution, action, structured evidence, and up to
five compatible candidate summaries for possible merging.

The output uses the following structure; descriptive strings below denote
the requested fields rather than a logged model response.

\begin{quote}
\small
\begin{verbatim}
{
  "event_relevant": true,
  "hypothesis": {
    "condition": "when this policy applies",
    "policy": "a semantic principle without concrete element IDs",
    "action_type": "the observed action type",
    "expected_effect": [
      {"field": "url_changed", "op": "eq", "value": true}
    ],
    "failure_evidence": [
      {"field": "error_detected", "op": "eq", "value": true}
    ],
    "confidence": 0.0,
    "merge_target_id": "an existing candidate ID or empty string"
  }
}
\end{verbatim}
\end{quote}

Predicates use the operators \texttt{eq}, \texttt{ne}, \texttt{gt},
\texttt{ge}, \texttt{lt}, and \texttt{le}, with a JSON scalar as the comparison
value. Permitted evidence fields include reward, positive reward, URL and
page-type changes, content and result changes, goal progress, state novelty,
action loops, errors, terminal status, the next page type and website, and
action type. The prompt requires falsifiable effects and prohibits inferring
generality or efficiency from a single transition.

A proposed merge target must be an existing candidate with compatible scope,
action type, and effects. The implementation checks these conditions and
does not permit merging into verified items. Merging preserves the stable
identifier and records aliases and source episodes; repeated discovery does
not increment validation support.

\section{Training Configurations}
\label{app:webarena-training}

\subsection{WebArena-Lite}

\paragraph{Supervised fine-tuning.}
The WebArena SFT baseline starts from Qwen3.5-35B-A3B and uses 9,415 samples
converted from the WebRL SFT data. Conversion maps the original actions to
BrowserGym commands, removes model-specific special tokens, and supplies the
actor system prompt. Unsupported quote actions are excluded. These are
external WebRL demonstrations rather than trajectories collected in the
reported evaluation stream.

Training updates all parameters using ZeRO-3 on eight H20 GPUs. The learning
rate is $10^{-5}$ with a cosine schedule and 3\% warmup, over three epochs.
The configuration uses a per-device batch size of one, four gradient
accumulation steps, and a sequence limit of 16,384 tokens, with BF16,
FlashAttention 2, Liger kernels, and gradient checkpointing. Evaluation uses
the final checkpoint without validation-based checkpoint selection.

\paragraph{Adaptation of TTRL.}
The WebArena training baseline adapts TTRL to the browser environment using
rejection sampling in place of the original policy-optimization procedure.
It samples eight rollouts per task with temperature 1.0, thinking disabled,
and a limit of 25 actions. Final answers are normalized and clustered by
exact matching or text similarity at a threshold of 0.85. Voting selects
pseudo-labels; an environment-success filter additionally excludes unsuitable
training trajectories. Thus, this implementation uses environment feedback
in data filtering, even though it does not use that feedback as the voted
pseudo-label.

The stream updates after batches of 32 tasks, with a replay fraction of 0.15.
Each update trains LoRA adapters of rank 16 and scale 32 on all target modules,
using a learning rate of $5\times10^{-5}$, one epoch, and a sequence limit of
12,288 tokens on four H20 GPUs. Adapters are merged before serving the updated
model. A fixed nine-task probe determines whether an update is retained or
rolled back. The final WebArena pipeline comprises two passes with ten
updates per pass, all retained.

\subsection{ALFWorld}

\paragraph{Supervised fine-tuning.}
The ALFWorld SFT baseline uses trajectories generated by the base model
during its first round of interaction. Both successful and unsuccessful
trajectories are retained, and the replay harness reconstructs the recorded
interactions byte for byte. Training updates all parameters on four GPUs
for one epoch. The resulting self-R1 checkpoint is evaluated with frozen
parameters, thinking enabled, two few-shot demonstrations, and the VA50
valid-action configuration. It achieves 392 successes in 670 episodes
(58.5\%).

\paragraph{Streaming TTRL.}
The ALFWorld TTRL baseline starts from the self-R1 SFT checkpoint rather
than directly from the base model. The streaming configuration uses
\texttt{n=16}, an update interval of 16 tasks, two passes, and 15\% replay.
A 20-task probe provides a circuit breaker for updates.
The final checkpoint, \texttt{model\_ttrl\_stream\_p1\_v6}, is evaluated
with frozen parameters under the same VA50, two-shot, and thinking settings.
This evaluation uses two rounds per task, yielding 197 successes in
268 episodes (73.5\%). These evaluation episodes are distinct from the
updates in the training stream.

\section{Training Cost Breakdown}
\label{app:training-cost}

Table~\ref{tab:training-hours} gives the GPU hours used in the training-cost
totals. GPU hours are computed from recorded durations and GPU counts.
At \$2.0 per H20 GPU-hour, the SFT and TTRL totals are \$84.0 and \$184.0,
respectively. The TTRL entries include rollout collection and parameter
updates in the retained pipelines; final local evaluation and discarded
development runs are excluded.

\begin{table}[htbp]
\centering
\caption{Training resources and costs for the retained pipelines.}
\label{tab:training-hours}
\small
\setlength{\tabcolsep}{10pt}
\begin{tabular}{@{}llrr@{}}
\toprule
Method & Benchmark & H20 GPU-hours & Cost (USD) \\
\midrule
SFT & WebArena-Lite & 29.8 & 59.6 \\
SFT & ALFWorld & 12.2 & 24.4 \\
\midrule
TTRL & WebArena-Lite & 38.0 & 76.0 \\
TTRL & ALFWorld & 54.0 & 108.0 \\
\bottomrule
\end{tabular}
\end{table}

\section{A Recorded Knowledge Example}
\label{app:knowledge-example}

A WebArena shopping interaction illustrates the distinction between the
discovery evidence and the proposed effect. In task 46, run 2, the actor
selects \texttt{click('227')} to open the \emph{My Account} link while seeking
the user's completed orders. The resulting URL changes from the shopping
homepage to \texttt{/customer/account/}; page type changes from a list to an
account repository, with no detected error and an immediate reward of zero.
The learner records the following hypothesis:

\begin{quote}
\textbf{Condition:} On a shopping homepage or list page, when opening the
customer account area through the My Account link.\par
\textbf{Policy:} Click the account link to navigate to the customer account
repository page.\par
\textbf{Expected effect:} \texttt{url\_changed = true} and
\texttt{goal\_progress = true}.\par
\textbf{Failure evidence:} \texttt{error\_detected = true}.
\end{quote}

The concrete element identifier belongs to the observed action and is omitted
from the reusable policy. The source transition supplies direct navigation
evidence, while goal progress is a prediction for later evaluation. Subsequent
records mark the item as verified and contain explicit usage attribution in
at least eight episodes involving account and order queries. The source
episode ends after three steps; the available excerpt does not establish
runtime use of this particular item before that episode ends. The example
therefore illustrates acquisition and later reported use, without treating
runtime eligibility as an observed use event.

\end{document}